\documentclass{article}
\pdfoutput=1
\usepackage{amsmath,amsthm,amssymb,graphicx,float,authblk}
\usepackage{slashbox}
\newtheorem{assumption_confusion_matrix}{Assumption}
\newtheorem{theorem_estimator}{Theorem}

\title{Model enhancement and personalization using weakly supervised learning for multi-modal mobile sensing}

\author[1]{Diyan Teng}
\author[1]{Rashmi Kulkarni}
\author[1]{Justin McGloin}
\affil[1]{Qualcomm Techonologies Inc.}

\date{}

\begin{document}
\maketitle

\begin{abstract}
Always-on sensing of mobile device user's contextual information is critical to many intelligent use cases nowadays such as healthcare, drive assistance, voice UI. State-of-the-art approaches for predicting user context have proved the value to leverage multiple sensing modalities for better accuracy. However, those context inference algorithms that run on application processor nowadays tend to drain heavy amount of power, making them not suitable for an always-on implementation. We claim that not every sensing modality is suitable to be activated all the time and it remains challenging to build an inference engine using power friendly sensing modalities. Meanwhile, due to the diverse population, we find it challenging to learn a context inference model that generalizes well, with limited training data, especially when only using always-on low power sensors. In this work, we propose an approach to leverage the opportunistically-on counterparts in device to improve the always-on prediction model, leading to a personalized solution. We model this problem using a weakly supervised learning framework and provide both theoretical and experimental results to validate our design. The proposed framework achieves satisfying result in the IMU based activity recognition application we considered.
\end{abstract}
%
%
\section{Introduction}
\label{sec:1}
As the rapid growth of computational capability of mobile chips, more and more augmented intelligent (AI) applications start to deviate from cloud based solutions to explore on device methods. The trend is not a surprise as mobile device users gain better understanding of data privacy~\cite{zhao2018privacy,osia2017hybrid} and prefer high reliability, low latency experience~\cite{varghese2016challenges,li2018learning,lane2015early}. Meanwhile, many research works~\cite{zhu2015mining,wang2012context} have indicated that the performance of on device AI applications can be enhanced if correct contextual information about the user can be leveraged. For example, speaker recognition~\cite{bisio2018smart} can localize to use different model if user context shows him/her in a meeting room or in a bus. GPS can be pre-activated and smartphone message can be disabled when the context shows the user starts to drive~\cite{park2017automatic}. Gesture/speech recognition~\cite{heck2016location,grubert2016towards} can lead to different intention given user's location context shows in car, in bedroom, or in a restaurant.

Nevertheless, the contextual information would become worthless if the cost of computation is way higher than the value it brings. In practice, the cost maps to both the infrastructure cost which include sensor price, battery consumption, component size, and the algorithm cost which include runtime complexity, memory requirement, training data gathering effort. For example, GPS is an informative source of information widely used in the recognition of mode of transportation~\cite{feng2013transportation}. However, the on time power consumption when GPS is in tracking mode is typically at the order of 10's mA, assuming one second EPOCH. This is discouraging designers from including GPS in an always-on transportation mode recognition engine in mobile devices. More importantly, we observe that the context information, typically, does not vary frequently in time. Thus, always-on low power solutions that leverage simple sensors and adopt low complexity algorithm are preferred. 

Even though, it seems a trivial solution to enable high cost sensing modalities in always-on inference, we claim that they can still be leveraged in a opportunistic fashion to maximize the system efficiency. Currently, we are exploring two approaches to incorporate the information opportunistically. On one hand, we can use those information as a confirmation or correction to the always-on prediction result. On the other hand, we consider leveraging the information to perform on device learning to personalize and enhance the  always-on inference model for each particular user, however, only if they are available. In this work, we focus on the second route. Specifically, we model the problem as a weakly supervised learning task, where the annotation is provided opportunistically through the high cost, non-always-on sensors. In mobile sensing it can lead to significant improvement if the inference model is personalized to the target. In other words, it turns out to be extremely challenging in most of the mobile sensing use cases to build a model with good interpersonal generalization performance. And this has become the primary motivation of this work.

The rest of this work is organized as follows. In Section~\ref{sec:2} we review related arts on weakly supervised learning to serve as a background. In Section~\ref{sec:3} we propose our main algorithm that leverage the ideas in weakly supervised learning to learn the target distribution of interest. In Section~\ref{sec:4}, we show that the proposed algorithm is statistically consistent and provide some insight that relates the performance of the algorithm to the noise statistics in the annotation. In Section~\ref{sec:5}, we provide both synthetic example and an application to validate our theory. Finally, we conclude and point out a few problems that worth further research.

\section{Multimodal weakly supervised learning}
\label{sec:2}

The definition of \emph{weakly} supervised learning varies in literature. To the best of our knowledge, state-of-the-art works have focused on three types of weakly supervised learning problems. The first scenario considers that only part of the data are labeled. This is also known as semi--supervised learning in some works. The objective in this case is to leverage prior knowledge, such as geometry of the data, to optimize prediction power on the labeled instances while encouraging the prior knowledge to be satisfied on both labeled and unlabeled instances. The second scenario is known as the positive--unlabeled (PU) learning. Under this category, only part of the instances from the positive hypothesis are labeled. The challenge is to properly handle the negative instances, without explicit knowledge of the label, so that the algorithm does not overfit due to the label imbalance. The last set of problem can be categorized as learning with label noise. In other words, even though labeled, the training instances may not be perfectly supervised. Instead, some noise adding process is considered so that the learner has no direct access to the ground truth. There can be many practical reasons for this to happen. For example, privacy requirement and adversarial attack.

In this paper, we consider the last type of problem, however, under a slightly different setup. Instead of directly assuming some label noise behavior, we consider the label to be obtained through a separate inference process, which is imperfect. The following graphical model illustrates the concept we are discussing about:

\begin{figure}[H]
\centering
\includegraphics[width=0.5\columnwidth]{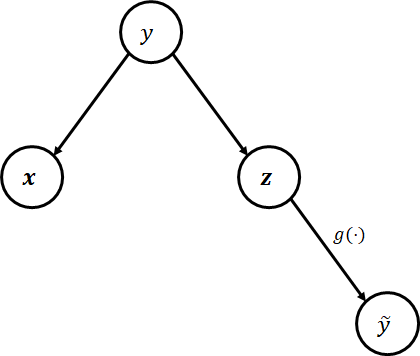}
\caption{Graphical model for weakly supervised framework}
\label{graphical_model}
\end{figure}

Here, $y \in \{s_1,s_2,\cdots,s_K\}$ denotes the ground truth class label which takes value in a discrete finite alphabets. $\mathbf{x}$ and $\mathbf{z}$ represent two independent measurements which contain class conditional information about $y$. The objective here is to learn the statistical relationship between $\mathbf{x}$ and $y$. In this work, we consider to learn the generative distribution $p(\mathbf{x}|y)$ and assume the prior distribution $p(y)$ is given. However, we are not directly given the pair $\{\mathbf{x}_n,y_n\}_{n=1}^N$. Instead, there is an inference process $g(\cdot)$ which takes $\mathbf{z}$ as input and predicts $y$ as $\tilde{y}$. In other words, we have no access to the generative distribution of $p(\mathbf{x}|y)$ and $p(\mathbf{z}|y)$, but only the pair $\{\mathbf{x}_n,\tilde{y}_n\}_{n=1}^N$. We start by assuming the predictive model $g(\cdot)$ is trained separately beforehand and fixed. Moreover, the experts who trained the model $g(\cdot)$ can also provide their confusion statistics $\pmb \Pi$ that understands their performance for predicting $y$ using $\tilde{y}$. Specifically, the $(i,j)$th element of the confusion matrix $\pmb \Pi$ represents the probability $\Pr(y=s_i|\tilde{y}=s_j)$.\footnote{Alternatively, one can also define the forward confusion matrix which represents $\Pr(\tilde{y}=s_i|y=s_j)$.} Later, we will consider $g(\cdot)$ can be flexible and discuss its effect on the learning process.

From a practical system design perspective, this model captures a few common use cases in mobile sensing. For example, modality--$\mathbf{x}$ can be heavily user dependent such as speech, face image or motion kinetics while modality--$\mathbf{z}$ can be invariant to user identity such as speed of traveling on traffic, altitude change or illumination level \emph{et al}. Therefore, we may enhance and personalize the prediction model for modality--$\mathbf{x}$. Also, the price for obtaining label $\tilde{y}$ might itself be different. For example, we can always ask the user to provide an annotation, which is the most accurate but expensive. In comparison, if we design an annotator using modality--$\mathbf{z}$ that runs in background, it will not interfere with the user experience. But, the label will become imperfect as a consequence. It is also worth mentioning that typically the predictive power using modality--$\mathbf{z}$ is worse than using modality--$\mathbf{x}$ or tends to be more power hungry. Because, otherwise the problem statement becomes trivial, and there will be no value to improve the model for $\mathbf{x}$.

\section{Noise correction estimator}
\label{sec:3}
In this section, we describe our algorithm for recovering the generative distribution of $p(\mathbf{x}|y)$ with access only to the pairs $\{\mathbf{x}_n,\tilde{y}_n\}_{n=1}^N$. We start with the following assumptions on the confusion matrix $\pmb \Pi$. 

\begin{assumption_confusion_matrix}\label{assumption1}
The confusion matrix $\pmb \Pi$ is a proper left stochastic (Markov) matrix. \emph{i.e.} $\sum_{i} \pmb \Pi_{i,j} = 1$.
\end{assumption_confusion_matrix}

\begin{assumption_confusion_matrix}\label{assumption2}
The confusion matrix $\pmb \Pi$ is invertible. \emph{i.e.} $\text{det}(\pmb \Pi) \neq 0$.
\end{assumption_confusion_matrix}

\begin{assumption_confusion_matrix}\label{assumption3}
The inference algorithm $g(\cdot): \mathbf{z}\to \tilde{y}$ is deterministic.
\end{assumption_confusion_matrix}

\begin{figure}[H]
\centering
\includegraphics[width=0.5\columnwidth]{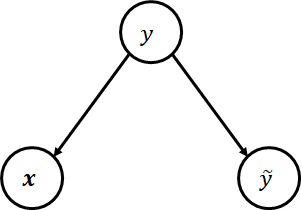}
\caption{Graphical model using deterministic inference rule for modality--$\mathbf{z}$}
\label{graphical_model_simplified}
\end{figure}

The first assumption can be easily met. Typically the confusion matrix is provided by the designer of the inference system using modality--$\mathbf{z}$ through empirical evaluation on some validation dataset. Therefore, it requires only proper column normalization. Later, we will understand the second assumption ensures the recovery of the generative distribution $p(\mathbf{x}|y)$ to be feasible. Intuitively, if the confusion matrix is not full rank, then information of some alphabet in $y$ will be lost in modality--$\mathbf{z}$. Therefore, it will become not recoverable. Moreover, we assumed the inference algorithm $g(\cdot)$ to be deterministic, which will most like be the case in practical system for complexity concern. Therefore, we can simplify the graphical model to Figure~\ref{graphical_model_simplified}.

From Figure~\ref{graphical_model_simplified} we observe that since $y$ is unobserved, the generative distribution of $p(\mathbf{x}|\tilde{y})$ can be written as:
\begin{equation}
\begin{split}
p(\mathbf{x}|\tilde{y}) & = \sum_{y} p(\mathbf{x},y|\tilde{y})\\
& = \sum_{y} p(\mathbf{x}|y) p(y|\tilde{y})
\end{split}
\end{equation}
If we write it in matrix format, we have:
\begin{equation}
\begin{split}
&
\begin{bmatrix}
p(\mathbf{x}|\tilde{y}=s_1)&p(\mathbf{x}|\tilde{y}=s_2)&\cdots&p(\mathbf{x}|\tilde{y}=s_K)
\end{bmatrix}\\
=
&
\begin{bmatrix}
p(\mathbf{x}|y=s_1)&p(\mathbf{x}|y=s_2)&\cdots&p(\mathbf{x}|y=s_K)
\end{bmatrix}
\cdot \pmb \Pi
\end{split}
\end{equation}
Finally, the generative distribution of $p(\mathbf{x}|y)$ can be recovered by inverting the confusion matrix:
\begin{equation}
\begin{split}
&
\begin{bmatrix}
p(\mathbf{x}|y=s_1)&p(\mathbf{x}|y=s_2)&\cdots&p(\mathbf{x}|y=s_K)
\end{bmatrix}
\\
=
&
\begin{bmatrix}
p(\mathbf{x}|\tilde{y}=s_1)&p(\mathbf{x}|\tilde{y}=s_2)&\cdots&p(\mathbf{x}|\tilde{y}=s_K)
\end{bmatrix}
\cdot \pmb \Pi^{-1}
\end{split}
\end{equation}
In practice, we also have no access to the true distribution of $p(\mathbf{x}|\tilde{y})$. And it has to be learned from the stored training instances $\{\mathbf{x}_n,\tilde{y}_n\}_{n=1}^N$. Denote the estimator of $p(\mathbf{x}|\tilde{y})$ as $q(\mathbf{x}|\tilde{y})$, we can similarly calculate the noise corrected estimator of $p(\mathbf{x}|y)$ as:
\begin{equation}\label{estimator_likelihoodFun}
\begin{split}
&
\begin{bmatrix}
q(\mathbf{x}|y=s_1)&q(\mathbf{x}|y=s_2)&\cdots&q(\mathbf{x}|y=s_K)
\end{bmatrix}
\\
=
&
\begin{bmatrix}
q(\mathbf{x}|\tilde{y}=s_1)&q(\mathbf{x}|\tilde{y}=s_2)&\cdots&q(\mathbf{x}|\tilde{y}=s_K)
\end{bmatrix}
\cdot \pmb \Pi^{-1}
\end{split}
\end{equation}
which simplifies to
\begin{equation}\label{estimator_likelihoodFunSimp}
q(\mathbf{x}|y=s_i) = \sum_{j} q(\mathbf{x}|\tilde{y}=s_j)\cdot \pmb \Pi_{(j,i)}^{-1}
\end{equation}
However, we need to be aware that the estimator~\eqref{estimator_likelihoodFunSimp} may not be a proper probability distribution. Even though, it automatically satisfy the condition that the integral over $\mathbf{x}$ equals one, there can be regions where this function take negative values. Therefore, it becomes challenging how to construct $q(\mathbf{x}|\tilde{y})$ when $\mathbf{x}$ is a continuous random vector.
In the next section, we will prove that this noise correction estimator is indeed lossless when number of stored training instances goes to infinity. Similarly, the posterior probability $p(y|\mathbf{x})$ can be estimated by following this method. We have, for the posterior probability:
\begin{align*}
p(\tilde{y}|\mathbf{x}) & = \sum_{y} p(y,\tilde{y}|\mathbf{x})\\
& = \sum_{y} p(y|\mathbf{x})p(\tilde{y}|y,\mathbf{x})\\
& = \sum_{y} p(y|\mathbf{x})p(\tilde{y}|y)
\end{align*}
Thus we have:
\begin{equation*}
\begin{split}
&
\begin{bmatrix}
p(y=s_1|\mathbf{x})&p(y=s_2|\mathbf{x})&\cdots&p(y=s_K|\mathbf{x})
\end{bmatrix}
\\
=
&
\begin{bmatrix}
p(\tilde{y}=s_1|\mathbf{x})&p(\tilde{y}=s_2|\mathbf{x})&\cdots&p(\tilde{y}=s_K|\mathbf{x})
\end{bmatrix}
\cdot \pmb \Pi_{\mathrm{R}}^{-1}
\end{split}
\end{equation*}
where $\Pi_{\mathrm{R}}$ in this case represents the right stochastic matrix whose $(i,j)$th element measure $\Pr(\tilde{y}=s_j|y=s_i)$.

Here, we also observe that estimator~\eqref{estimator_likelihoodFunSimp} is interestingly related to the method of unbiased estimator proposed by Natarajan \emph{et al.}~\cite{natarajan2013learning}. Specifically, they considered a binary classification problem where $y \in \{+1,-1\}$ with the following label flipping probabilities:
\begin{equation}
\kappa_{+1} = \Pr(\tilde{y}=+1|y=-1),\quad \kappa_{-1} = \Pr(\tilde{y}=-1|y=+1)
\end{equation}
satisfying the constraint:
\begin{equation}
\kappa_{+1} + \kappa_{-1} < 1
\end{equation}
They proposed an unbiased estimator for the loss function that can be adopted in an empirical risk minimization (ERM) procedure as:
\begin{equation}
\tilde{l}(f(\mathbf{x}),y) = \frac{(1-\kappa_{-y})l(f(\mathbf{x}),y)-\kappa_{y}l(f(\mathbf{x}),-y))}{1-\kappa_{+1}-\kappa_{-1}}
\end{equation}
Their result can indeed be understood as the frequentists' counterpart of the probabilistic framework we proposed here. Additionally, we can generalize the theory to a multiclass setting and construct:
\begin{equation}
\begin{split}
\pmb K \cdot
\begin{bmatrix}
\tilde{l}(f(\mathbf{x}),\tilde{y}=s_1)&\cdots&\tilde{l}(f(\mathbf{x}),\tilde{y}=s_K)
\end{bmatrix}^{\text{T}} \\
=
\begin{bmatrix}
l(f(\mathbf{x}),y=s_1)&\cdots&l(f(\mathbf{x}),y=s_K)
\end{bmatrix}^{\text{T}}
\end{split}
\end{equation}
where $\pmb K$ in this case is the right stochastic matrix with $\pmb K_{(i,j)}=\Pr(\tilde{y}=s_i|y=s_j)$. Similarly, we assume the forward confusion matrix $\pmb K$ to be invertible, \emph{i.e.} $\text{det}(\pmb K)\neq 0$. Therefore, the multiclass unbiased risk can be calculated as:
\begin{equation}\label{multiclass_ubrisk}
\begin{split}
\begin{bmatrix}
\tilde{l}(f(\mathbf{x}),\tilde{y}=s_1)&\cdots&\tilde{l}(f(\mathbf{x}),\tilde{y}=s_K)
\end{bmatrix}^{\text{T}} \\
= \pmb K^{-1} \cdot
\begin{bmatrix}
l(f(\mathbf{x}),y=s_1)&\cdots&l(f(\mathbf{x}),y=s_K)
\end{bmatrix}^{\text{T}}
\end{split}
\end{equation}
which simplifies to
\begin{equation}
\tilde{l}(f(\mathbf{x}),\tilde{y}=s_i) = \sum_{j} {\pmb K_{(i,j)}^{-1}} l(f(\mathbf{x}),y=s_j)
\end{equation}
One interesting observation here is that the original requirement $\kappa_{+1} + \kappa_{-1} < 1$ given in~\cite{} appears to be not necessary. Instead, in the binary case, one can easily prove that the invertibility condition translates into only $\kappa_{+1} + \kappa_{-1} \neq 1$. This matches the properties of the receiver operating characteristics (ROC) curve in decision theory~\cite{}. In principle, ROC curve is always above the straight line passing through $\mathrm{P_D}=\mathrm{P_{FA}}=0$ and $\mathrm{P_D}=\mathrm{P_{FA}}=1$, which corresponds to the Bernoulli random guess decision. Otherwise, the decision rule can be flipped to achieve that. In~\eqref{multiclass_ubrisk}, the matrix inverse operation can automatically handle the decision flipping procedure when $\kappa_{+1} + \kappa_{-1} > 1$.

\section{Consistency}
\label{sec:4}
We analyze the consistency of our estimator defined in~\eqref{estimator_likelihoodFun} in this section. The key challenge in~\eqref{estimator_likelihoodFun} is to analyze the effect of the inverse stochastic matrix $\pmb \Pi^{-1}$ on the density estimator. We start by establishing a theory that guarantees estimator~\eqref{estimator_likelihoodFun} can recover the true generative distribution given certain conditions.

\begin{theorem_estimator}\label{thm1}
Let $D_{\text{KL}}^{(i)}(p_i \| q_i)$ be the Kullback--Leibler (KL) divergence that measures the discrepancy between the two distributions $p(\mathbf{x}|y=s_i)$ and $q(\mathbf{x}|y=s_i)$. Specifically,
\begin{equation}
D_{\text{KL}}^{(i)}(p_i \| q_i) = \int p(\mathbf{x}|y=s_i) \log \frac{p(\mathbf{x}|y=s_i)}{q(\mathbf{x}|y=s_i)} \mathrm{d} \mathbf{x}
\end{equation}
Similarly, let $\tilde{D}_{\text{KL}}^{(i)}(\tilde{p}_i \| \tilde{q}_i)$ denote the KL divergence between $p(\mathbf{x}|\tilde{y}=s_i)$ and $q(\mathbf{x}|\tilde{y}=s_i)$. Suppose both $q(\mathbf{x}|y)$ and $q(\mathbf{x}|\tilde{y})$ are valid distribution functions and Assumptions 1--3 hold, we have
\begin{equation}
\sum_{i} D_{\text{KL}}^{(i)}(p_i \| q_i) = 0
\end{equation}
if and only if
\begin{equation}
\sum_{i} \tilde{D}_{\text{KL}}^{(i)}(\tilde{p}_i \| \tilde{q}_i) = 0
\end{equation}
\end{theorem_estimator}
\begin{proof}
To prove the necessity part, observe that
\begin{align*}
\sum_{i} \tilde{D}_{\text{KL}}^{(i)}(\tilde{p}_i \| \tilde{q}_i) = \sum_{i} \int p(\mathbf{x}|\tilde{y}=s_i) \log \frac{p(\mathbf{x}|\tilde{y}=s_i)}{q(\mathbf{x}|\tilde{y}=s_i)} \mathrm{d} \mathbf{x} \\
= \sum_{i} \int \underbrace{\sum_{j} p(\mathbf{x}|y=s_j) \pmb \Pi_{(j,i)} \log \frac{\sum_{j} p(\mathbf{x}|y=s_j) \pmb \Pi_{(j,i)}}{\sum_{j} q(\mathbf{x}|y=s_j) \pmb \Pi_{(j,i)}}}_{(\ast)} \mathrm{d} \mathbf{x}
\end{align*}
Apply log--sum inequality to $\ast$ yields
\begin{align*}
(\ast) \leq \sum_{j} p(\mathbf{x}|y=s_j) \pmb \Pi_{(j,i)} \log \frac{p(\mathbf{x}|y=s_j) \pmb \Pi_{(j,i)}}{q(\mathbf{x}|y=s_j) \pmb \Pi_{(j,i)}}\\
= \sum_{j} p(\mathbf{x}|y=s_j) \pmb \Pi_{(j,i)} \log \frac{p(\mathbf{x}|y=s_j)}{q(\mathbf{x}|y=s_j)}
\end{align*}
Thus
\begin{align*}
& \sum_{i} \tilde{D}_{\text{KL}}^{(i)}(\tilde{p}_i \| \tilde{q}_i)\\
\leq & \sum_{i,j} \pmb \Pi_{(j,i)} \int p(\mathbf{x}|y=s_j) \log \frac{p(\mathbf{x}|y=s_j)}{q(\mathbf{x}|y=s_j)} \mathrm{d} \mathbf{x}\\
= & \sum_{i,j} \pmb \Pi_{(j,i)} D_{\text{KL}}^{(j)}(p_j \| q_j)
\end{align*}
Necessity follows since if $D_{\text{KL}}^{(i)}(p_i \| q_i) = 0$ for all $i$, then $\tilde{D}_{\text{KL}}^{(i)}(\tilde{p}_i \| \tilde{q}_i) = 0$ for all $i$.

For sufficiency, the log--sum property cannot be used since $\pmb \Pi_{(j,i)}^{-1}$ may be negative valued. Instead, we evaluate
\begin{align*}
&{D}_{\text{KL}}^{(i)}({p}_i \| {q}_i)\\
= & \int \sum_{j} p(\mathbf{x}|\tilde{y} = s_j) \pmb \Pi_{(j,i)}^{-1} \log \frac{\sum_{j} p(\mathbf{x}|\tilde{y} = s_j) \pmb \Pi_{(j,i)}^{-1}}{\sum_{j} q(\mathbf{x}|\tilde{y} = s_j) \pmb \Pi_{(j,i)}^{-1}} \mathrm{d} \mathbf{x}
\end{align*}
by directly observing that since $p(\mathbf{x}|\tilde{y}=s_j) \overset{a.e.}{=} q(\mathbf{x}|\tilde{y}=s_j)$ implies $p(\mathbf{x}|\tilde{y}=s_j) \pmb \Pi_{(j,i)}^{-1} \overset{a.e.}{=} q(\mathbf{x}|\tilde{y}=s_j) \pmb \Pi_{(j,i)}^{-1}$ for finite $\pmb \Pi^{-1}$. Therefore, the integral remains to be zero.
\end{proof}

Theorem~\ref{thm1} guarantees that when sample size goes to infinity, a perfect estimator for $\tilde{p}$ yields a perfect estimator for $p$. Next, we discuss how the confusion matrix affect the convergence rate.

\begin{theorem_estimator}\label{thm2}
Let $\mathbb{G}_n^{(i)}$ denotes the following empirical process
\begin{equation}
\begin{split}
& \mathbb{G}_n^{(i)} = \int \left( \log q_i - \log p_i \right) \cdot \mathrm{d} (\mathbf{\bar{P}}^{(i)} -\mathbf{\bar{P}}_n^{(i)})\\
& = \int \left( \log \sum_j \tilde{q}_j \pmb \Pi_{(j,i)}^{-1} - \log \sum_j \tilde{p}_j \pmb \Pi_{(j,i)}^{-1} \right)\\
& \cdot \mathrm{d} (\mathbf{\bar{P}}^{(i)} -\mathbf{\bar{P}}_n^{(i)})
\end{split}
\end{equation}
where $\mathbf{\bar{P}^{(i)}}$ denotes the true distribution of $\mathbf{x}|y=s_i$ and $\mathbf{\bar{P}}_n^{(i)}$ is the empirical version of it.
Then we have the following condition holds for $\mathbb{G}_n^{(i)}$:
\begin{equation}
\mathbb{G}_n^{(i)} = \mathcal{O}(\log \frac{\lambda_{\text{max}}}{\lambda_{\text{min}}}\cdot n^{-\frac{1}{2}} \vee \max_j \mathbb{\tilde{G}}_n^{(j)}) 
\end{equation}
where $\lambda_{\text{max}}$ and $\lambda_{\text{min}}$ are respectively the maximal and minimal eigenvalue of $\pmb \Pi^{-1} \cdot \pmb \Pi^{-1 \mathrm{T}}$. And 
\begin{equation*}
\mathbb{\tilde{G}}_n^{(j)} = \int \left( \log \tilde{q}_j - \log \tilde{p}_j \right) \mathrm{d} (\mathbf{\tilde{P}}^{(j)} - \mathbf{\tilde{P}}_n^{(j)})
\end{equation*}
denotes the empirical process that measures the convergence of individual density estimator.
\end{theorem_estimator}
\begin{proof}
Rewrite $\mathbb{\tilde{G}}_n^{(j)}$ as
\begin{align*}
&\mathbb{\tilde{G}}_n^{(j)} = \int \log \frac{\sum_j \tilde{q}_j \pmb \Pi_{(j,i)}^{-1}}{\sum_j \tilde{p}_j \pmb \Pi_{(j,i)}^{-1}} \mathrm{d} (\mathbf{\bar{P}}^{(i)} -\mathbf{\bar{P}}_n^{(i)})\\
&= \frac{1}{2} \int \log \frac{ (\sum_j \tilde{q}_j \pmb \Pi_{(j,i)}^{-1})^2 }{ (\sum_j \tilde{p}_j \pmb \Pi_{(j,i)}^{-1})^2 } \mathrm{d} (\mathbf{\bar{P}}^{(i)} -\mathbf{\bar{P}}_n^{(i)})\\
&= \frac{1}{2} \int \log \frac{\mathbf{\tilde{q}}^{\mathrm{T}} \pmb \Pi_{(:,i)}^{-1} \pmb \Pi_{(:,i)}^{-1 \mathrm{T}} \mathbf{\tilde{q}}}{\mathbf{\tilde{p}}^{\mathrm{T}} \pmb \Pi_{(:,i)}^{-1} \pmb \Pi_{(:,i)}^{-1 \mathrm{T}} \mathbf{\tilde{p}}} \mathrm{d} (\mathbf{\bar{P}}^{(i)} -\mathbf{\bar{P}}_n^{(i)})\\
&\leq \frac{1}{2} \int \log \frac{\lambda_{\text{max}}}{\lambda_{\text{min}}} \cdot \frac{ \|\mathbf{\tilde{q}}\|^2 }{ \|\mathbf{\tilde{p}}\|^2 } \mathrm{d} (\mathbf{\bar{P}}^{(i)} -\mathbf{\bar{P}}_n^{(i)})\\
&= \frac{1}{2} \int \log \frac{\lambda_{\text{max}}}{\lambda_{\text{min}}} \mathrm{d} (\mathbf{\bar{P}}^{(i)} -\mathbf{\bar{P}}_n^{(i)}) + \frac{1}{2} \int \log \frac{ \|\mathbf{\tilde{q}}\|^2 }{ \|\mathbf{\tilde{p}}\|^2 } \mathrm{d} (\mathbf{\bar{P}}^{(i)} -\mathbf{\bar{P}}_n^{(i)})
\end{align*}
where the inequality is based on the fact that the eigenvalues of $\pmb \Pi$ satisfies $0 < \frac{1}{\sqrt{|\lambda_i|}} \leq 1$. For the second term, we observe the rate for $\sum_{j} \tilde{q}_j^2$ to converge to $\sum_{j} \tilde{p}_j^2$ is determined by the slowest term since each individual term is strictly positive. Therefore, the upper bound to it is $\max_j \mathbb{\tilde{G}}_n^{(j)}$. For the first term, from central limit theorem we have $\int \log \frac{\lambda_{\text{max}}}{\lambda_{\text{min}}} \mathrm{d} (\mathbf{\bar{P}}^{(i)} -\mathbf{\bar{P}}_n^{(i)}) = \mathcal{O}(\log \frac{\lambda_{\text{max}}}{\lambda_{\text{min}}}\cdot n^{-\frac{1}{2}})$.
\end{proof}

Theorem~\ref{thm2} provides insight on how the confusion process affect the learning rate. Specifically, in addition to the learning rate of each individual density estimator, an extra cost has to be paid based on how much information is lost during the confusion process. To give one example, when the confusion matrix is any permutation matrix, there will be no loss of information but only deterministic label swapping are performed. Thus, we have $\lambda_{\text{max}}=\lambda_{\text{min}}=1$ and no additional cost will be paid. In contrast, if we have a close to singular confusion matrix, the loss of information will be high, because there are multiple $\tilde{y}$ now representing highly similar information in $y$. In this case, $\lambda_{\text{max}}$ is large. 

Finally, for the risk function constructed in~\eqref{multiclass_ubrisk}, it can be proven to be unbiased.

\begin{theorem_estimator}
The risk function estimator defined in~\eqref{multiclass_ubrisk} is an unbiased estimator of $l(f(\mathbf{x},y))$:
\begin{equation}
\mathrm{E}_{\tilde{y}} [\tilde{l}(f(\mathbf{x},\tilde{y}))] = l(\mathbf{x},y)
\end{equation}
\end{theorem_estimator}
\begin{proof}
For every $y = s_i$, we have
\begin{equation}\label{risk_expectation}
\mathrm{E}_{\tilde{y}|y=s_i}[\tilde{l}(f(\mathbf{x},\tilde{y}))] = \sum_{\tilde{y}} \Pr(\tilde{y}|y=s_i) \tilde{l}(f(\mathbf{x},\tilde{y}))
\end{equation}
To have~\eqref{risk_expectation} to be an unbiased estimator of $l(f(\mathbf{x},y=s_i))$ for every $s_i$, we need the following equalities to hold
\begin{equation}
\begin{split}
\pmb K \cdot
\begin{bmatrix}
\tilde{l}(f(\mathbf{x}),\tilde{y}=s_1)&\cdots&\tilde{l}(f(\mathbf{x}),\tilde{y}=s_K)
\end{bmatrix}^{\text{T}} \\
=
\begin{bmatrix}
l(f(\mathbf{x}),y=s_1)&\cdots&l(f(\mathbf{x}),y=s_K)
\end{bmatrix}^{\text{T}}
\end{split}
\end{equation}
which gives~\eqref{multiclass_ubrisk} when $\mathbf{K}$ is invertible.
\end{proof}

\section{Experimental results}
\label{sec:5}
In this section, we provide synthetic example and real world applications for our theory. We start by considering a synthetic example. We draw binomial samples from three classes with success parameters $[0.52,0.65,0.08]$ respectively. The class annotations are corrupted using a confusion matrix $\pmb \Pi$. We select the confusion matrix to have identical diagonal components and identical off-diagonal components to simplify the experiment. And the noise level in this case can be controlled by adjusting the off-diagonal elements while maintaining the rows sums to one. We evaluate the performance of estimator~\eqref{estimator_likelihoodFunSimp} by computing the sum of KL divergence between the estimated distribution and the true distribution, using the analytic form.

\begin{figure}[H]
\centering
\includegraphics[width=\linewidth]{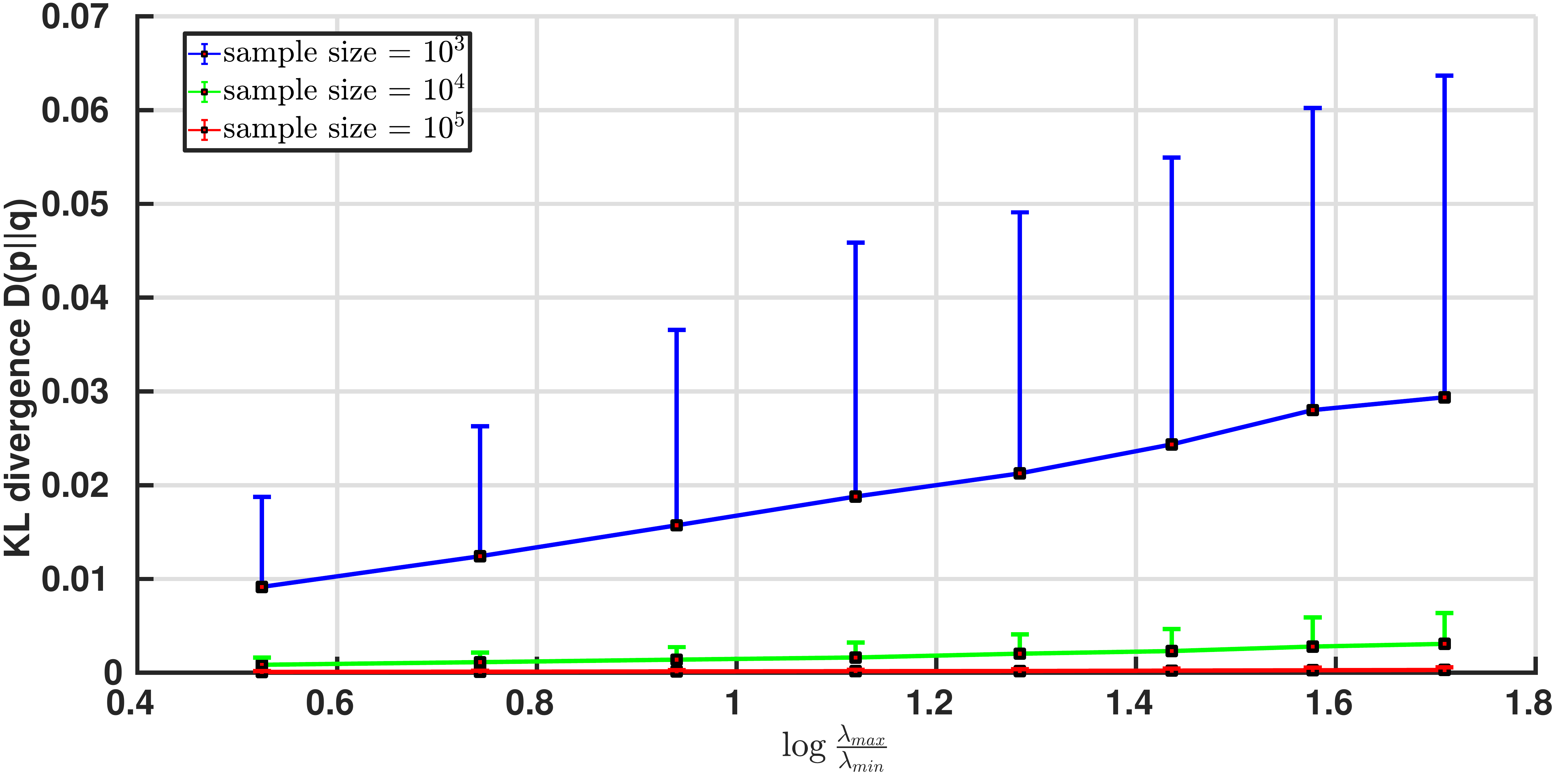}\\
\includegraphics[width=\linewidth]{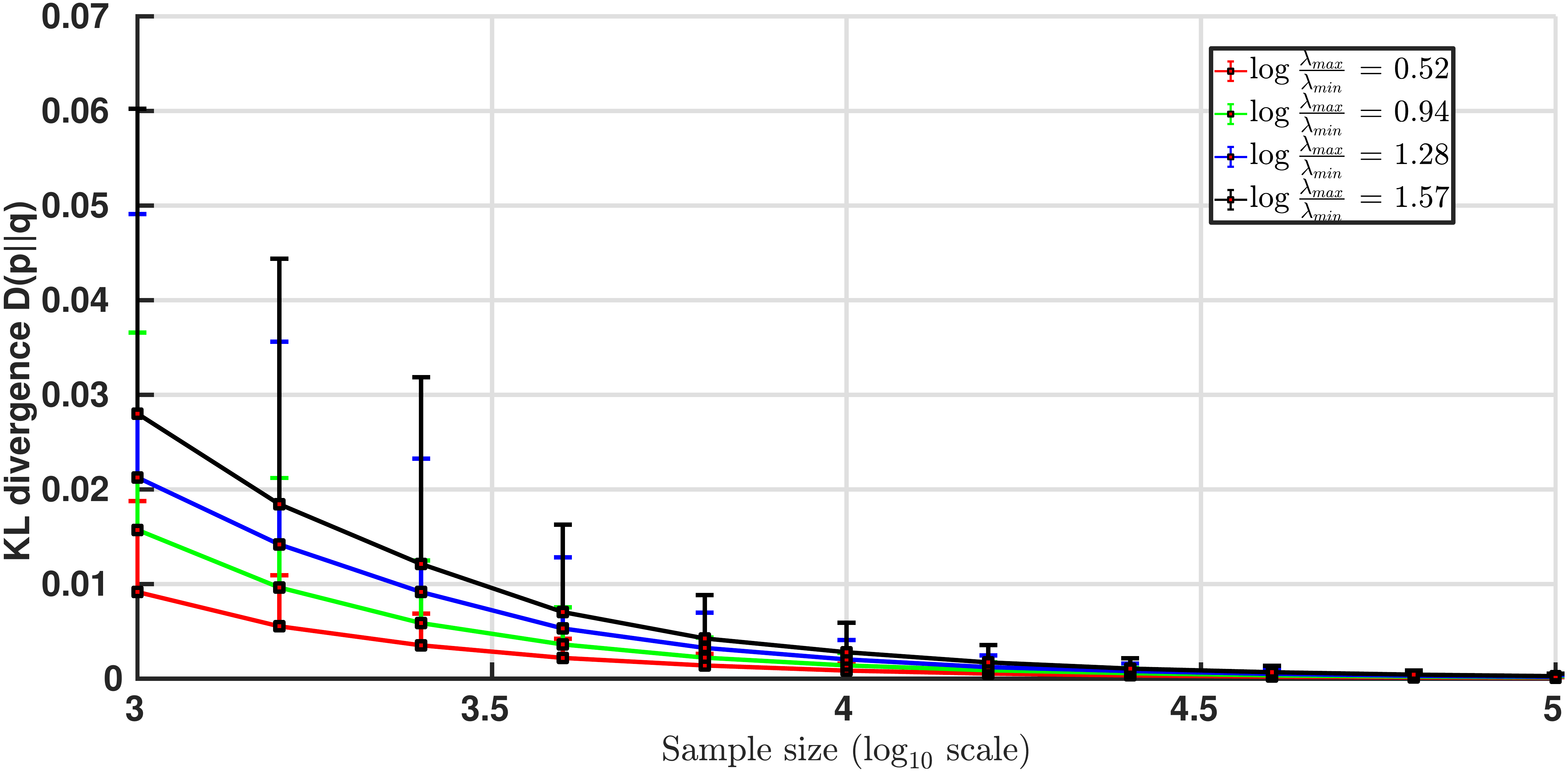}\\
\includegraphics[width=\linewidth]{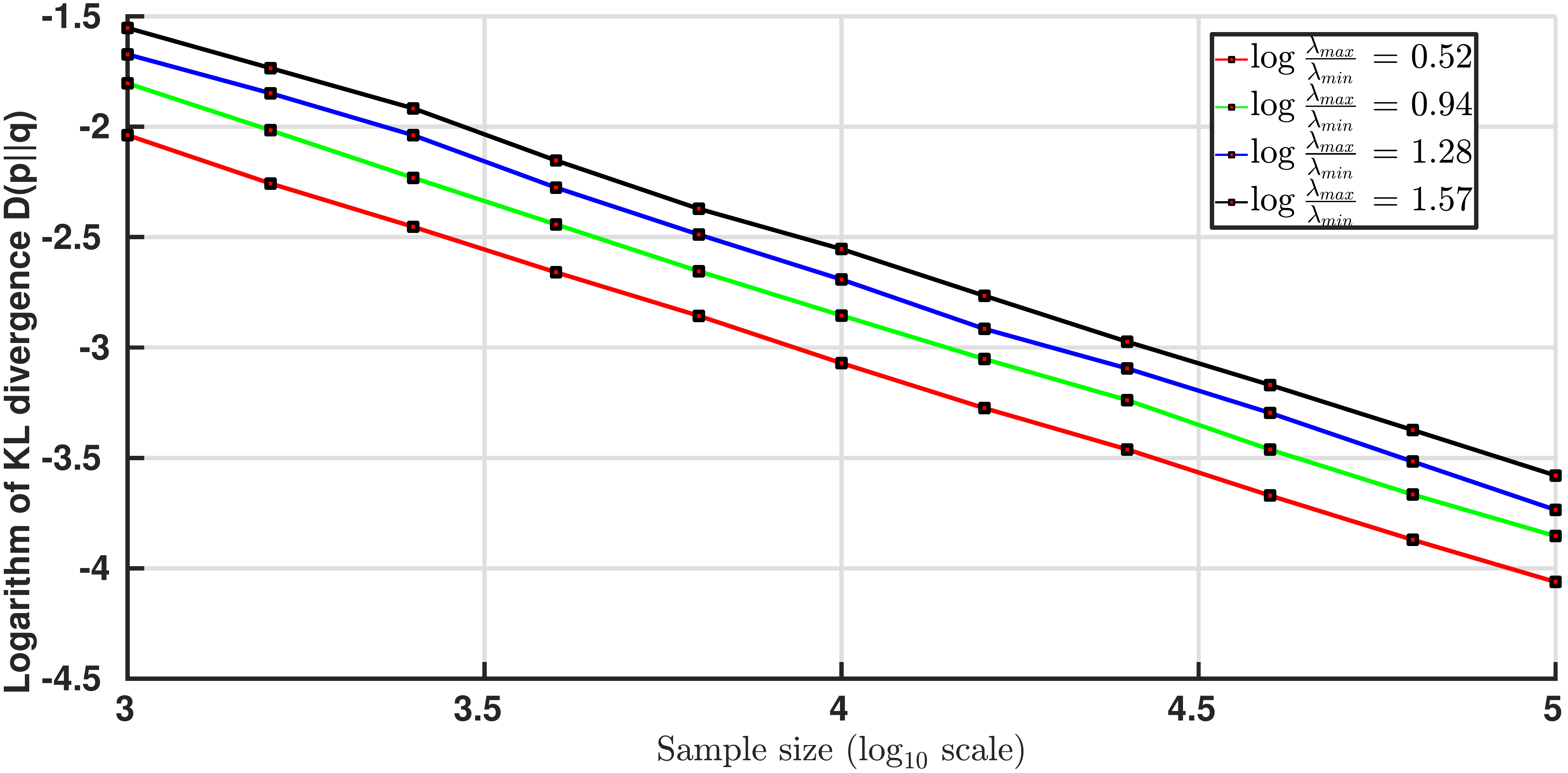}
\caption{Convergence in sum of KL divergence: (a) as $\frac{\lambda_{\text{max}}}{\lambda_\text{min}}$ increases; (b) as sample size increases; (c) Figure (b) in logarithm scale. Solid square indicates the mean over 2000 test runs. Error bar show the one sided standard deviation.}
\end{figure}

We observe the convergence behavior in terms of $\log \frac{\lambda_{\text{max}}}{\lambda_{\text{min}}}$ is close to linear for each fixed sample size. And the convergence behavior in sample size in this example can be theoretically proven to be $\mathcal{O}(n^{-\frac{1}{2}})$ using central limit theorem.

\begin{figure}[H]
\centering
\includegraphics[width=\linewidth]{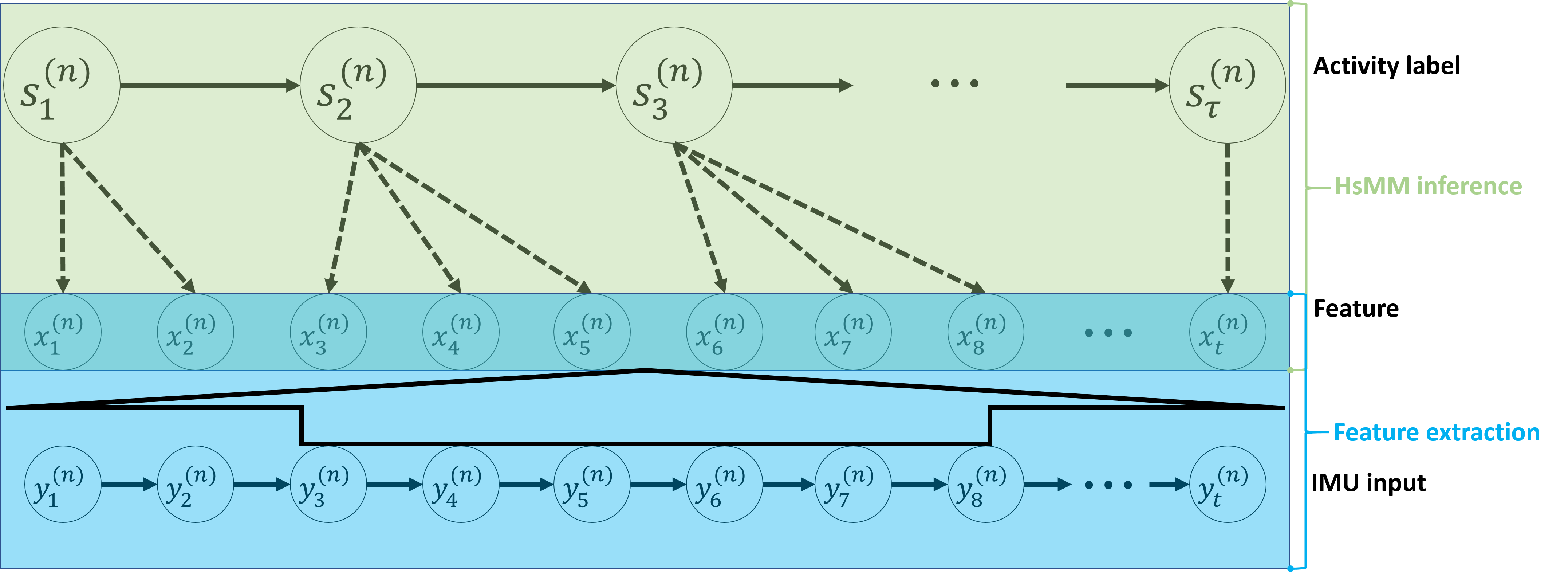}
\caption{Graphical model for activity recognition using HsMM}
\label{HsMM_AR}
\end{figure}

Next we apply the proposed algorithm in a real application. We consider activity recognition using smartphone accelerometer and gyroscope sensors. As we noticed for many users, it is a challenging task to distinguish phone call\footnote{Holding phone close to ear while speaking. People tends to pace very slowly around without moving to a particular destination.} and slow walk ($\leq 0.3$Hz step rate). Also, for some users, their slow walk and biking signatures can be hard to distinguish. Thus, we select these three classes and apply our algorithm. Our basic classifier is a Bayes network that converts the time series input from sensors to activity class probabilities. We built an hierarchical model that first extract features from the time series and then perform smoothed prediction using a hidden semi-Markov model (HsMM). The graphical model is shown in Figure~\ref{HsMM_AR}. The feature extraction layer convert the time series into a set of finite alphabets, however, we cannot reveal details about the feature extraction block due to confidential reason. We explore next if the inference layer can be re-trained and enhanced by leveraging GPS speed readings. As GPS readings are power consuming, they are not always available. In addition, we cross validated that for these three classes, a simple threshold based classifier would leads to the confusion statistics in Table~\ref{gps_conf}.

\begin{table}[H]
\footnotesize
\centering
  \begin{tabular}{| c | c | c | c |}
  \hline
  \backslashbox{Truth}{Result} & call (fidget) & slow walk & bike\\ \hline
  call (fidget) & 0.76 & 0.24 & 0\\ \hline
  slow walk & 0.28 & 0.72 & 0\\ \hline
  bike & 0 & 0 & 1\\ \hline
  \end{tabular}
  \caption{Confusion statistics using GPS speed readings. Fidget: $[0, 0.1]$mph. Slow walk: $(0.1, 1]$mph. Bike: $(3,25]$mph. No readings for these three classes are within $(1,3]$mph.}
  \label{gps_conf}
\end{table}

\begin{figure}[H]
\centering
\includegraphics[width=\linewidth]{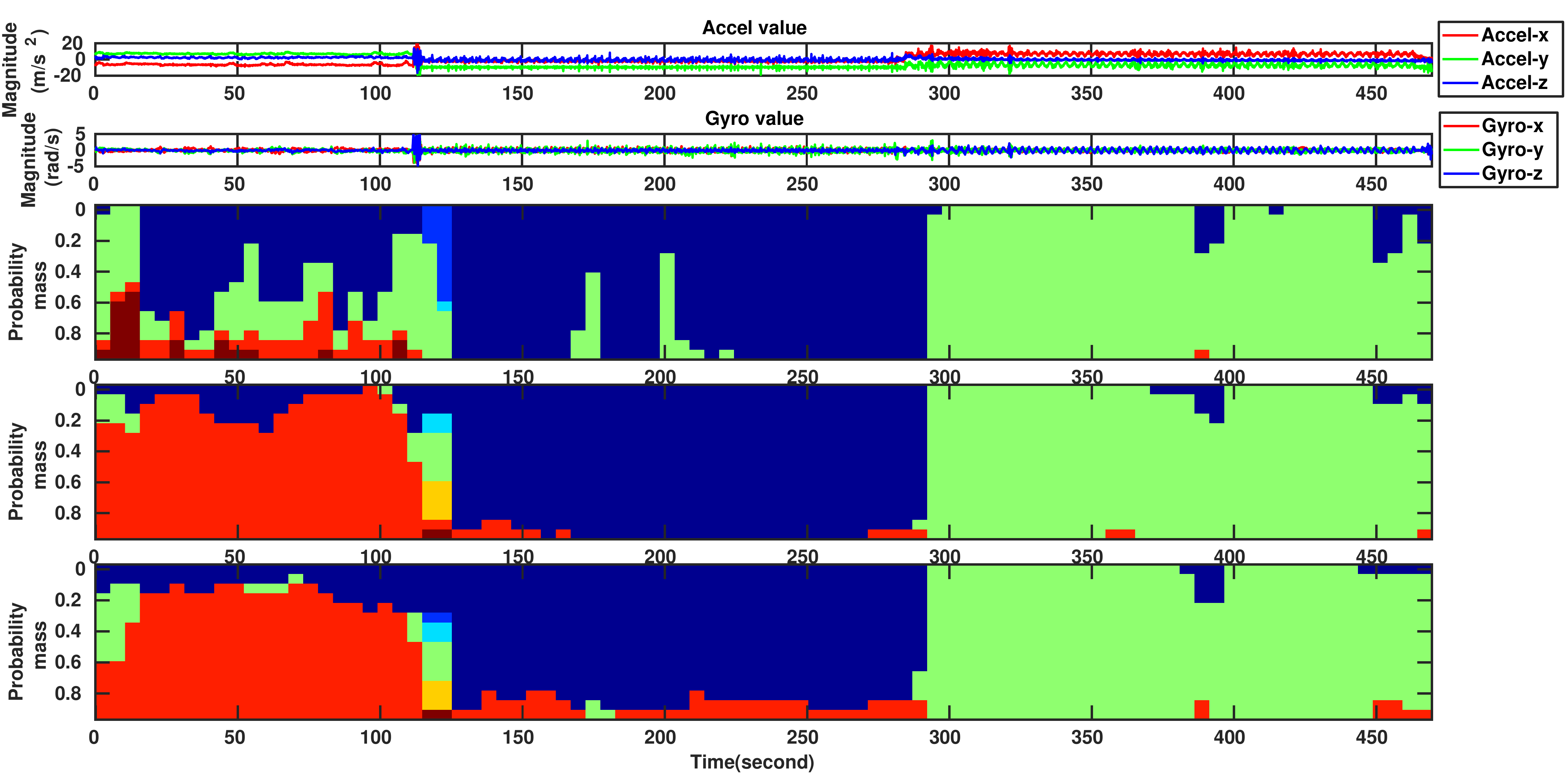}\\
\includegraphics[width=\linewidth]{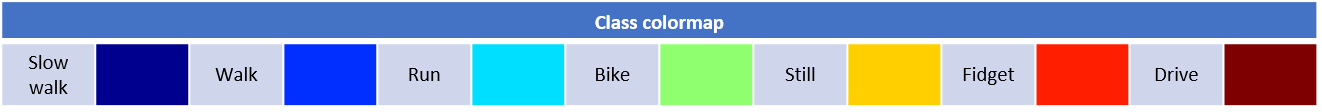}
\caption{Activity recognition model personalization experiment. Row 1--2: accelerometer and gyroscope signal. (Order of events: call $\to$ slow walk $\to$ bike) Row 3: inference using baseline model. Row 4: inference using personalized model (ground truth annotation). Row 5: inference using personalized model (GPS based annotation). Probability mass is computed by marginalizing stochastic sequence realizations (samples from the HsMM posterior distribution) at each time slot.}
\label{prob_mass_model_comparison}
\end{figure}

The testing users are required to collect two minutes of data within each category for re-training purpose followed by an uninterrupted collection of transition among those three classes. The baseline model is trained on our company internal dataset which contain not only these three classes but a few other classes. Next, the baseline model is personalized using the two minutes clean collection but annotated in two different ways. The first annotation is to use the ground truth. And the second annotation is to use the GPS speed based classifier in Table~\ref{gps_conf}. Then estimator~\eqref{estimator_likelihoodFunSimp} is used to correct the annotation noise. We select a Dirichlet--Multinomial pair for the HsMM's emission probability, where the Dirichlet prior parameters are set according to the posterior values of the baseline model. After re-training, those parameters are updated again using the clean data. The baseline model, personalized model using ground truth and personalized model using GPS inputs are tested on the transition data for comparison in Figure~\ref{prob_mass_model_comparison}.

As we may observe, the baseline model is not correctly recognizing call as fidget. Instead, it creates some confusion between slow walk and bike. Subsequently, after we personalize the HsMM emission model using the separate collection paired with ground truth annotation, the model has gained significant confidence to correctly recognize call as a fidget event. Finally, the model personalized with GPS based annotation also achieves a satisfying recognition result. A detailed empirical Bayes error rate (BER) in this experiment is provided in Table~\ref{empirical_BER}.

\begin{table}[H]
\footnotesize
\centering
  \begin{tabular}{| c | c | c | c |}
  \hline
  Model type & Mdl1 & Mdl2 & Mdl3\\ \hline
  BER & 0.22 & 0.04 & 0.05\\ \hline
  \end{tabular}
  \caption{Empirical BER comparison. Mdl1: Baseline model. Mdl2: Personalized model (ground truth annotation). Mdl3: Personalized model (GPS based annotation).}
  \label{empirical_BER}
\end{table}

\section{Conclusion}
In this work, we proposed an automated annotation method for personalization of always-on mobile sensing model. The proposed algorithm leverages the non-always-on sensing modalities opportunistically. Synthetic results show that our algorithm can find the correct generative model given enough data. Our application shows the model can indeed help to improve smartphone based human activity recognition performance in some cases.

Nevertheless, some problems remain open. First, it is still challenging to construct and verify the generative model estimated, whether it satisfy basic probability measure properties, especially for high dimensional and continuous random variables. Second, as we noticed that the convergence rate is governed by both sample size and the eigenvalue structure of the confusion matrix, it is worth investigating if some tradeoff can be defined to perform sample selection. For example, if in addition to the noisy annotation, we are also provided a confidence measure for that annotation, it is interesting to consider subsampling the data for re-training. As rejecting samples with low confidence can leads to cleaner confusion statistics, it reduces the amount of samples that are available to learn the generative distribution. Also, in situation where training needs to happen on edge, it is important for mobile devices to save as less data as possible due to storage constraints.

\bibliographystyle{IEEEtran} 
\bibliography{wsl_cite}

\end{document}